%% file: 0_Main_Compiled.tex
\documentclass[letterpaper, 10 pt, conference]{ieeeconf}
\usepackage{PackageHeading}
\usepackage[absolute,overlay]{textpos}

\title{\Large \bf
Resource-Constrained Station-Keeping for Helium Balloons\\ using Reinforcement Learning
}

\author{Jack Saunders$^{1, 2}$, Loïc Prenevost$^{2}$, Özgür \c{S}im\c{s}ek$^{1}$, Alan Hunter$^{3}$, and Wenbin Li$^{1}$% <-this % stops a space
\thanks{This work is supported by Lux Aerobot and the UKRI Centre for Doctoral Training in Accountable, Responsible \& Transparent AI (ART-AI), under UKRI grant number EP/S023437/1.}% <-this % stops a space
\thanks{$^{1}$Department of Computer Science,
        University of Bath, UK, \{\protect\url{js3442, os435, w.li}\}\protect\url{@bath.ac.uk}}
\thanks{
        $^{2}$Lux Aerobot, Montréal Canada, \protect\url{lprenevost@luxaerobot.com}}
\thanks{
        $^{3}$Department of Mechanical Engineering,
        University of Bath, UK, \protect\url{A.J.Hunter@bath.ac.uk}}%
}

\begin{document}

\begin{textblock*}{12cm}(4.75cm,1cm) % {block width} (coords)
   \centering
   \textcolor{red}{
   This work has been submitted to the IEEE International Conference on Intelligent Robots and Systems (IROS) for possible publication. Copyright may be transferred without
notice, after which this version may no longer be accessible.}
\end{textblock*}

\maketitle
\thispagestyle{empty}
\pagestyle{empty}

\begin{abstract}
High altitude balloons have proved useful for ecological aerial surveys, atmospheric monitoring, and communication relays.  However, due to weight and power constraints, there is a need to investigate alternate modes of propulsion to navigate in the stratosphere.  Very recently, reinforcement learning has been proposed as a control scheme to maintain the balloon in the region of a fixed location, facilitated through diverse opposing wind-fields at different altitudes.  Although air-pump based station keeping has been explored, there is no research on the control problem for venting and ballasting actuated balloons, which is commonly used as a low-cost alternative.  We show how reinforcement learning can be used for this type of balloon. Specifically, we use the soft actor-critic algorithm, which on average is able to station-keep within 50\;km for 25\% of the flight, consistent with state-of-the-art.  Furthermore, we show that the proposed controller effectively minimises the consumption of resources, thereby supporting long duration flights.  We frame the controller as a continuous control reinforcement learning problem, which allows for a more diverse range of trajectories, as opposed to current state-of-the-art work, which uses discrete action spaces.  Furthermore, through continuous control, we can make use of larger ascent rates which are not possible using air-pumps.  The desired ascent-rate is decoupled into desired altitude and time-factor to provide a more transparent policy, compared to low-level control commands used in previous works.  Finally, by applying the equations of motion, we establish appropriate thresholds for venting and ballasting to prevent the agent from exploiting the environment.  More specifically, we ensure actions are physically feasible by enforcing constraints on venting and ballasting.
\end{abstract}

\subfile{1_Introduction.tex}

\subfile{2_Literature_Review.tex}

\subfile{3_Background.tex}

\subfile{4_Method.tex}

\subfile{5_Results.tex}

\subfile{6_Conclusion.tex}

\newpage
\bibliographystyle{ieeetr}
\bibliography{references.bib}

\end{document}

%% file: 1_Introduction.tex
\section{Introduction}
High altitude balloons are vital for researchers in areas such as environmental monitoring \cite{adams_continuous_2020, wang_optical_2021}, atmospheric analysis  \cite{walczak_light_2021}, and communication relay \cite{vandermeulen_distributed_2018, bellemare_autonomous_2020}.
% Disadvantages of using typical unmanned aerial vehicles
%                       Rephrase
Due to power and weight constraints, typical unmanned aerial vehicles and actively propelled balloons are infeasible.  Hence researchers have been investigating passive propulsion utilising wind-vectors at various altitudes to propel the balloon towards the target.
% The difficulty of passive propulsion
% Use rl to overcome non-linear relationshio
% use sim
The difficulty with this approach is the complex non-linear relationship between actions performed on the balloon and optimal states which lead the balloon towards the target.  This is further compounded by the wind-forecast uncertainty which cannot be accurately modeled.  As a result of the non-linear relationship, reinforcement learning (RL) has been adopted in place of search-based methods due to the problem's computational intractability.

% Previous work 
Previous work on RL-based station-keeping employ super-pressure balloons that depend on a high-strength plastic envelope to prevent the expansion of the lifting gas.  Envelopes made from high-strength plastic can be cost prohibitive for research applications.  In comparison, researchers have shown latex balloons can be made for under \$1000 \cite{sushko_low_2017} making them more accessible. Hence, we model a latex-balloon and show how RL can also be used for low-cost applications.

% Our contributions specifically
% Score
Our contributions are as follows:

% Limited Resource usage 
{(1)} To suit low-cost applications,  we introduce a model that accounts for limited resources, where the RL controller is trained to minimise resource usage while maximizing the time within the intended region. This is opposed to previous studies which model the diurnal cycle as a result of using renewable solar energy \cite{bellemare_autonomous_2020}
% Score
More specifically our RL controller achieves an average score of 25\% time flying within 50 km of the target for 3 days, consistent with the current state-of-the-art \cite{bellemare_autonomous_2020}.
% Continuous Control

{(2)} We frame the RL problem as a continuous control problem to take advantage of a range of ascent-rates, possible when using latex balloons, which are decomposed to desired altitude and time-factor which enables further transparency of the policy.
%Which differs from the prior literature's approach of employing discrete actions.   
% Decouple actions for transparency
%Furthermore, the decoupled actions of desired altitude and time-factor 
%we decouple the actions to provide more transparent trajectories which are easier to understand when compared to the wind-field.  
% Cap on minimum resources
To prevent the agent from exploiting the environment and incorporating practical minimum venting and ballasting times, we incorporate minimum thresholds.
% Published online

{(3)} Finally, the simulator is published  as an extension of Google's Balloon Learning Environment \cite{greaves_balloon_2021}, in the form of an Open-ai gym environment to allow other researchers to train RL algorithms on a venting/ballasting balloon.\footnote{More information on this environment can be found at \url{https://sites.google.com/view/resource-constrained-balloon}}

%% file: 2_Literature_Review.tex
\section{Literature Review}
% Purpose -> say how a limiting factor of this research is the cost associated to flying balloons.  Latex is the cost effective method.
\subsection{Applications}

% Aerial surveys to collect wildlife - better than drones as they can stay in the air for longer.
High altitude balloons provide a powerful means of collecting ecological data \cite{adams_continuous_2020, wang_optical_2021} as opposed to other unmanned aerial vehicles.  This is due to the extended and noise-reduced flight characteristics of balloons, which serve to limit the potential for disrupting wildlife.  
% Nocturnal light and atmospheric emmisions, can measure events from a different perspective.
In addition to ecological data, nocturnal light emissions \cite{walczak_light_2021} and atmospheric aerosol formation \cite{laakso_hot-air_2007} have also been monitored using high altitude balloons.  The use of balloons as a measurement platform provides the opportunity to measure atmospheric effects or ground effects from an alternative perspective as opposed to ground-based measurements. 

% Communication Relays - Annother application formation control
High-altitude balloons are also showing promise as an alternative mode of high-speed wireless internet access to regions with inadequate internet infrastructure \cite{vandermeulen_distributed_2018}.  This is a challenge since traversing through the stratosphere is a difficult task especially in strong wind-fields, as propulsion methods are severely restricted by power and weight limitations. 
% Definition of station-keeping
Station-keeping is a term used to describe the maneuvers used to maintain a vehicle's position, in this case, a balloon, relative to a target position.  While there are  obvious benefits for the application of communication relay, as previous mentioned, the ability to station keep has numerous other applications.

% From this go into propulsion...

\subsection{Propulsion}
To overcome strong wind-fields, researches have investigated a range of propulsion methods. 
%Horizontal Tandem
Lateral aerodynamic forces can help steer the balloon using stratosails \cite{ramesh_numerical_2018} or to reduce horizontal displacement using tandem tethered balloons \cite{zhang_tandem_2020}.  Alternatively, powered methods such as electro-hydrodynamic thrusters \cite{van_wynsberghe_station-keeping_2016} or propeller-based methods \cite{wang_recovery_2021} have been utilised for navigation.  Propeller-based methods, in particular, have been shown to be useful for recovering station-keeping missions. However, both electro-hydrodynamic and propeller-based methods are limited by the energy requirement and power constraints of the balloon. In addition, electro-hydrodynamic thrusters are limited by the scarcity of atmospheric particles at higher altitudes.  

%Horizontal Tandem
%Researchers have conducted extensive research into a diverse range of propulsion systems for navigating through the atmosphere.  Tandem tethered balloons can be used to exploit opposing wind velocities at different altitudes to reduce horizontal displacement \cite{zhang_tandem_2020}.  Similarly, a stratosail tethered onto the main balloon can provide lateral aerodynamic forces to help steer the balloon towards a desired position \cite{ramesh_numerical_2018}.

%Power required, battery constraints
%Electro-hydrodynamic thrusters use an electric field to accelerate ionized ambient air particles.  The movement of ions creates a flow and generates a thrust force \cite{van_wynsberghe_station-keeping_2016}.  The design of electro-hydrodynamic thrusters is characterized by the absence of moving parts, rendering them lightweight, reliable, and easy to maintain. However, the implementation of these thrusters is challenged by the scarcity of atmospheric particles at higher altitudes, owing to the decrease in air density.

%Finally, researchers have also investigated propeller-based propulsive systems to recover station-keeping missions \cite{wang_recovery_2021}.  The primary challenge in utilising propellers is the energy requirement from continuously rotating the blades.
%%%%%%%%%%%%%%%%%%%%%%%%%%%%

% Different methods of altitude propulsion
Alternatively, high altitude balloons can navigate through the atmosphere by taking advantage of changes in wind direction at different altitudes.  To do so, the main technique is to adjust the buoyancy or gravitational forces acting on the balloon.  One method is to heat the air inside the balloon envelope such that the air inside the balloon is less dense causing a pressure difference \cite{edmonds_hot_2009}. 
Another mode of altitude control uses air ballasts, typically used for superpressure balloons, which modulates the weight of the balloon by pumping air into and out of a pressurized container \cite{hall_altitude-controlled_2019}.  The most notable shortcoming of this approach is the ``parasitic mass'' added by the air ballast, which becomes increasingly massive with decreasing altitude therefore consuming more energy \cite{voss_altitude_2003}. 
Solar power has been shown effective to power the air-pump for longer flight times \cite{bellemare_autonomous_2020}.  Finally, latex balloons which vent lift gas and deploy preloaded ballasts have been shown to be an energy-efficient way to adjust the balloon's altitude \cite{sushko_low_2017}.  The use of venting and ballasting this way significantly reduces the energy needs of the balloon system.  Furthermore, a wide range of ascent rates can be achieved within a short space of time, depending on the valve size or ballast drop rate.

% Hot-air balloons
%Hot-air balloons use engines which are used to heat the air inside the envelope, causing the air to become less dense.  This density difference between the air outside and inside the balloon leads to a pressure difference and therefore the buoyancy force $Vg(\rho-\rho_i)$ \cite{edmonds_hot_2009}.
% Air-pump
%Alternatively air ballasts, or pumps, are used to modulate the weight of the balloon by pumping air into and out of a pressurized container \cite{hall_altitude-controlled_2019}.  Solar power can be used to power the pump for longer flight times \cite{bellemare_autonomous_2020}, however ascent rates are limited by the pump efficiency.
% Venting-ballast
%Venting lift gas and deploying preloaded ballasts has been shown to be an energy-efficient way to adjust the balloon's altitude \cite{sushko_low_2017}.  The use of venting and ballasting this way significantly reduces the energy needs of the balloon system.  Furthermore, a wide range of ascent rates can be achieved within a short space of time, depending on the valve size or ballast drop rate.
\subsection{Station-Keeping}
The primary factor contributing to the navigational uncertainty of a balloon is the availability of the desired wind speeds.  Researchers have investigated controllers to perform station keeping as a means of keeping the balloon within a radius of the target.  
%Geometric controller
Du {\it et al.} present a geometric approach to station keeping \cite{du_station-keeping_2019}.  When a balloon is beyond an intermediate radius, the desired altitude is determined by calculating the maximum angle between the orientation of the balloon relative to the target location and the direction of the wind, using the dot product.  This method assumes the wind fields are relatively stable which is not always the case.  
%RL controller Marc Bellemare
However, according to Bellemare {\it et al.}, standard model-predictive control tools face challenges in station-keeping due to the complex and non-linear relationship between control decisions and the distance to the target \cite{bellemare_autonomous_2020}.  They explain that wind uncertainty cannot be modeled in closed form and therefore the consequences of exploration cannot be quantified.  Furthermore, not only must the controller optimise over minimum distance from the target but also minimise the resources used.  This makes search-based controllers computationally impractical.  To overcome this issue, the authors use a variant of distributional RL known as Quantile Regression based Deep Q-learning \cite{dabney_distributional_2018}.  The controller is parameterised as a neural network and the output of the network represent control outputs for the air-pump. The authors model solar parameters to make use of renewable solar energy to power the pump.  The controller is then able to learn the effect of the diurnal cycle.  Their controller was trained on augmented ECMWF ERA5 \cite{hersbach_era5_2020} wind data and to overcome wind forecast errors the Gaussian Process was employed to calculate the uncertainty between the wind forecast and the true wind speed. 
% DQN from 
Very recently, Xu {\it et al.} use a deep Q-network to solve station keeping, showing how beneficial a prioritised experience replay based on high-value samples can improve the stability of training \cite{xu_station-keeping_2022}.  
%conclusion
While station-keeping using air-pumps has been explored earlier \cite{xu_station-keeping_2022, bellemare_autonomous_2020}, we show that RL can also be used for latex balloons that vent helium and ballast solid mass.  

%Furthermore, we frame the RL control problem as a limited resource problem and attempt to learn policies that minimise resources used, since the resources are not renewable.  Finally we make use of the increased ascent rate, achievable for venting/ballasting-based balloons by using the Soft Actor-Critic (SAC) algorithm.

%\textbf{Our contributions are the following}: (1) We present an RL based approach to perform station-keeping for high altitude balloons that are propelled using venting and ballasting. Our method is able to achieve on average $25$\% time within $50$\,km to the target.  (2) Then, due to resource constraints, we show how the RL agent learns how to conserve it's resources such that it can fly for a total of 3 days.  (3) We then publish the dynamic model, as an extension of Google's Balloon Learning Environment \cite{greaves_balloon_2021}, in the form of an Open-ai gym environment to allow other researchers to train RL algorithms on a venting/ballasting balloon.  More information on this environment can be found here \url{https://sites.google.com/view/resource-constrained-balloon}

%% file: 3_Background.tex
\section{Background}

\subsection{Equations of Motion}

The objective here is to derive an ordinary differential equation (ODE) for the ascent rate of the balloon as a function of time.  This ODE can then be solved given the initial state of the balloon.  The fundamental assumptions utilized to emulate the station-keeping performance of the balloon are outlined bellow:

\begin{enumerate}
    \item The variations of atmospheric temperature, pressure, and density with altitude are modeled using the US Standard Atmosphere Model $1976$.
    \item The helium inside the balloon envelope and atmospheric air are assumed to follow the ideal gas law.
    \item The helium gas inside the balloon is assumed to have uniform density, temperature, and pressure. This is because the balloon experiences rotations when moving through the wind field, which causes the gas to mix \cite{du_flight_2019, xu_station-keeping_2022}.
    \item Latex is composed of a stretchable polymer which expands as the balloon ascends.  Hence a very small pressure difference (approximately 150Pa) occurs between the helium and ambient air \cite{sushko_low_2017}.  This pressure difference is negligible at pressures less than 5000 Pa, Hence the internal pressure is assumed to be equivalent to the ambient pressure.
    \item The balloon reaches steady state before the controller performs another action.
    \item We assume the ambient temperature is equal to the internal temperature of the balloon.  Modelling this would enable us to model the diurnal cycle of the sun \cite{bellemare_autonomous_2020}, however, due to limitations in finding thermal radiation properties of latex, we decide to omit this as done in Sobester's study \cite{sobester_high-altitude_2014}.
\end{enumerate}

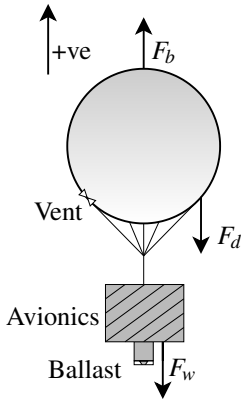
\begin{figure}[ht]
\centering

\begin{overpic}[]{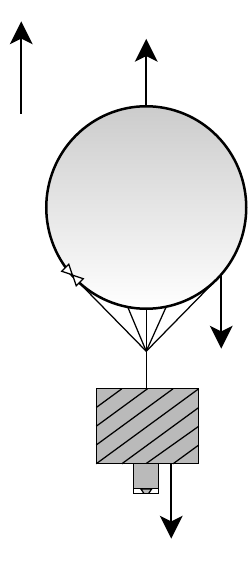}

\put(15, 140){\makebox(0,0){+ve}}
\put(50, 140){\makebox(0,0){$F_b$}}
\put(75, 70){\makebox(0,0){$F_d$}}
\put(57, 20){\makebox(0,0){$F_w$}}
\put(20, 20){\makebox(0,0){Ballast}}
\put(10, 80){\makebox(0,0){Vent}}
\put(8, 40){\makebox(0,0){Avionics}}
\end{overpic}

\caption{{\small Free body diagram, modelled as a point mass, illustrating the buoyancy $F_b$, drag $F_d$, and gravitational $F_w$ forces.}}
\label{fig:FreeBodyDiagram}
\end{figure}

We equate the sum of the forces acting on the balloon using the Newton-Euler formulation to calculate the {vertical motion} of the balloon.  The forces include buoyancy $F_b$, weight $F_w$, and drag $F_d$ which acts in the opposite direct of motion, as shown in Fig \ref{fig:FreeBodyDiagram}.

\begin{equation} \label{eq:N2L}
    m\Ddot{h} = F_b - F_d - F_w
\end{equation}

The buoyancy force $F_b=\rho_a V g$ is a result of the mass of the displaced air caused by a pressure differential within the air column.  Where the density of air $\rho$ and the volume of the balloon envelope $V$ vary with altitude.

The drag force $F_d = -\frac{1}{2}\rho c_d A |\dot{h}|\dot{h}$ represents the force acting opposite to the relative motion with respect to the surrounding fluid. 
Where $\dot{h}$ is the ascent rate, $A$ is the cross-sectional area and $c_d$ is the drag coefficient \cite{taylor_classical_2005}.  Du {\it et al.} use $c_d=0.2$ in their study \cite{du_station-keeping_2019}, citing Almedeji's study while assuming the wind around the balloon is turbulent, $Re>10^6$ \cite{almedeij_drag_2008}.  Whereas, Bellemare {\it et al.} use $c_d=0.25$ \cite{bellemare_autonomous_2020}.

Finally, the mass of the balloon can be divided into three distinct components: the mass of the payload $m_p$, which includes the balloon envelope, the mass of helium $m_h$, and the mass of sand ballast $m_s$. These components combine to form the total gravitational force acting upon the balloon, $F_w = (m_p + m_h + m_s)g$, where $g$ is the acceleration due to gravity.

We can now substitute these forces into Eq \ref{eq:N2L} and rearrange for $\Ddot{h}$.

\begin{equation}
\label{eq:fullN2L}
    \Ddot{h} = \frac{\left(\rho V g -\frac{1}{2}\rho c_dA
    \left|\dot{h}\right|\dot{h} - mg\right)}{m}
\end{equation}

%Using the ideal gas equation.
The volume occupied by the balloon envelope is dependent on the altitude which can be calculated using the ideal gas law $V=\frac{nRT}{P}$.  Where $P$ is the ambient pressure, $R$ is the universal gas constant, $n$ is the number of mols of helium, and $T$ is the ambient temperature.  It is assumed that the helium acts as an ideal gas and the internal temperature of the helium is equal to the ambient temperature.  Further work would include thermal radiation calculations and an experiment to calculate the reflection and absorption properties of the latex material.  Due to the elasticity properties of the latex envelope, the pressure of the helium is equal to the ambient pressure.  In practice, there is approximately $150$\,Pa of overpressure, which only becomes significant above $30$\,km \cite{sushko_low_2017}. 
Finally, adiabatic lapse rates are approximations used to calculate the temperature profile in the standard atmospheric model.  Given the temperature at a reference altitude $T_0$, the ambient temperature $T$ at a given altitude can be calculated by $T=T_0+(h \times L)$.  Then using the volume, calculated from the ideal gas law, we can calculate the radius of the balloon envelope and therefore the drag area $A=\pi r^2=\pi\left(\frac{3V}{4\pi}\right)^{\frac{2}{3}}$.  Finally, by rearranging the ideal gas law again, we can equate the density of air $\rho = \frac{PM_a}{RT}$, where $M_a$ is the molar mass of dry air.

Within the existing literature, two approaches have been proposed to model the {horizontal motion} of a balloon, which are based on either considering the drag force components \cite{du_station-keeping_2019, furfaro_wind-based_2008} or assuming that the balloon's horizontal velocity components are equal to the atmospheric wind components $v_{wx}, v_{wy}$ \cite{dai_performance_2012, morani_method_2009}. In our study, we utilize wind data obtained from the European Centre for Medium-Range Weather Forecasts (ECMWF) Reanalysis 5th generation (ERA5) dataset \cite{hersbach_era5_2020}, without a vertical component.  The use of the wind data-set precludes the calculation of the relative velocity of the wind with respect to the balloon. Thus, we adopt the latter assumption and assume that the balloon's horizontal velocity is equivalent to that of the wind.

Therefore, using state space form, we let $\vec{z} = [z_1, z_2, z_3, z_4, z_5, z_6]^T = [x, \dot{x}, y, \dot{y}, h, \dot{h}]^T$ represent our state vectors.
\begin{equation}
\begin{bmatrix}
    \dot{z_1} \\
    \dot{z_2} \\
    \dot{z_3} \\
    \dot{z_4} \\
    \dot{z_5} \\
    \dot{z_6} \\
\end{bmatrix}
=
\begin{bmatrix}
    v_{wx} \\
    0 \\
    v_{wy} \\
    0 \\
    z_6 \\
    \frac{\left(\rho V g -\frac{1}{2}\rho c_{drag}A
    \left|\dot{h}\right|\dot{h} - mg\right)}{m} \\
\end{bmatrix}
\end{equation}

\subsection{Soft Actor-Critic}

The reinforcement learning problem can be formulated as a Markov Decision Process (MDP), defined by a tuple $(\mathcal{S}, \mathcal{A}, P, R)$ \cite{sutton_reinforcement_2017}.  At each decision stage $t$, an agent in state $s_t\in\mathcal{S}$ takes an action $a_t\in\mathcal{A}$. Consequently, the agent receives reward $r_{t+1}$ according to reward distribution $R(s_t, a_t)$ and reaches a new state $s_{t+1}$ determined by the probability distribution $P(s_{t+1}|s_t,a_t)$.  The objective of the agent is to maximise the expected future cumulative reward $\mathbb{E}[G_t|s_t]$, where $G_t=\sum^T_{k=t+1}\gamma^{k-t-1}R_k$ and $\gamma\in[0,1]$ is the discount factor.   The state-action function can therefore defined as $Q_\pi(s,a)=\mathbb{E}_\pi[G_t|s_t, a_t]$.  Where the policy $\pi(a_t|s_t)$ is a mapping from states to probabilities of selecting each possible action.

Soft Actor-Critic \cite{haarnoja_off-policy_2018} is a an off-policy deep RL algorithm based on the maximum entropy framework.  Within this framework, the actor aims to maximise both the expected reward and the policy entropy (which measures the stochasticity of the policy).  SAC simultaneously learns a policy (actor) $\pi_\theta$ and a Q-function (critic) $Q_{\phi}$ which are parameterised using neural networks.  Since SAC is an off-policy learning algorithm, we can update both the actor and the critic using a replay buffer $\mathcal{D}$.  The Q-function parameters can be optimised using the mean squared loss function $J_Q(\theta)$, between the predicted and target action-value $Q_\phi$.  Where $-log(\pi(a_t|s_t))$ is the entropy term and $\alpha$ is the entropy temperature governing the weighting of the entropy.

\begin{align}
    J_Q(\phi)= & \mathbb{E}_{(s_t, a_t)\sim\mathcal{D}}\left[\frac{1}{2}\left(Q_\phi(s_t,a_t)-\hat{Q}(s_t,a_t)\right)^2\right] \\
\begin{split}
    \hat{Q}(s_t,a_t)= & r(s_t,a_t)+\gamma(1-d)\mathbb{E}_{a\sim\pi}[ Q_\phi(s_{t+1},\pi_\theta(a_{t+1}|s_{t+1})) \\ 
    & - \alpha\log\pi_\theta(a_{t+1}|s_{t+1})]
\end{split}
\end{align}

To optimise the policy, SAC utilises the reparameterisation trick, where a sample from $\pi_\theta$ is computed using a squashed Gaussian policy, thus the samples are obtained using $\tilde{a_t}_\theta(s, \epsilon)=\tanh(\mu_\theta(s)+\sigma_\theta(s)\odot\epsilon)$ where epsilon is sampled from a standard normal $\epsilon\sim\mathcal{N}(0, I)$.  The policy can then be optimized by minimising a simplified form of the Kullback-Leibler divergence using the objective function $J_\pi(\phi)$.

\begin{equation}
    J_\pi(\phi)=\mathbb{E}_{s_t\sim\mathcal{D},\epsilon_t\sim\mathcal{N}}[\log\pi_\phi(\tilde{a}(s_t, \epsilon_t)|s_t)-Q_\theta(s_t, \tilde{a}(s_t, \epsilon_t))]
\end{equation}

The entropy temperature $\alpha$ can be adjusted by taking the gradient of $J_\alpha$ \cite{haarnoja_soft_2019}:

\begin{equation}
    J_\alpha=\mathbb{E}[-\alpha\log\pi_\theta(a_t|s_t;a)-\alpha \bar{H}],
\end{equation}

where $\bar{H}$ is the desired minimum entropy, which for this study is set to 0.

%% file: 4_Method.tex
\section{Method}

\subsection{Wind Data}

Real wind forecasts are gathered from the  ECMWF's ERA5 global reanalysis dataset \cite{hersbach_era5_2020}.  As mentioned by Bellemare {\it et al.}, simplex noise \cite{perlin_image_1985} can be used to augment the wind data, thus emulating forecasting errors \cite{coy_global_2019}.  Specifically, wind forecasts between $1$st November $2022$ and $31$st January $2023$ are chosen as seasonal changes can have a big impact on the wind fields.  Furthermore, wind fields within the tropics, at longitude $-113^{\circ}$ latitude $1^{\circ}$, are chosen as they are shown to contain a range of diverse winds \cite{bellemare_autonomous_2020}.
% Had a stab at representing the wind field using set notation.  Will fix before submission.
The wind-field contains wind vectors over a grid of points in the parameter space $\mathcal{X} \times \mathcal{Y} \times \mathcal{P} \times  \mathcal{T}\times  \mathcal{V}$, where the longitude $\mathcal{X}$ and and latitudinal $\mathcal{Y}$ positions are sampled with a resolution of $0.4^{\circ}$ at various pressure points $\mathcal{P}$ ranging from $2000$\,Pa to $17500$\,Pa. Furthermore, forecasts are collected every $6$ hours for the three day episode and each wind-vector has two components in the longitude and latitude direction $(v_{wx}, v_{wy})$.

\subsection{High-level Soft Actor-Critic}

The Soft Actor-Critic algorithm is selected since the maximization of both the value function and the policy entropy is expected to assist in exploring the wind field, preventing the controller from getting trapped in local minima.

The MDP {state space} ($\mathcal{S}$) consists of $50$ wind and $27$ ambient features.  The wind features are collected at 25 equally-spaced points between the vertical pressure limits $[5000, 14000]$\,Pa.  For each point, the magnitude of the wind speed $|v|$ and bearing error $\theta$ with respect to the target is calculated, as in \cite{bellemare_autonomous_2020}.  These form two individual vectors which then are concatenated $[|v_0|,...,|v_{24}|,\theta_0,...,\theta_{24}]\in\mathbb{R}^{50}$.

The ambient features consist of simulated onboard measurements.  These measurements include the altitude $h_t$, ascent-rate $\dot{h}_t$, wind velocity $|v_h|$ and bearing error at the current altitude $\theta_h$, the balloon envelope drag area $A$ and volume $V$, total system mass $m_T$, distance $|x|$ and heading to the target $[\sin(\theta_x), \cos(\theta_x)]$.  The past three altitudes $[h_{t-1}, h_{t-2}, h_{t-3}]$, ascent rates $[\dot{h}_{t-1}, \dot{h}_{t-2}, \dot{h}_{t-3}]$, and float actions $[a_{2, t-1}, a_{2, t-2}, a_{2, t-3}]$.  Then finally the sand mass $m_s$, and helium mols $n$ are also included.  Both the wind and ambient features are then concatenated together to form a feature set of 74 variables.

We utilise the distance-based {reward function} proposed by Bellemare et al.  \cite{bellemare_autonomous_2020}, shown in Eq \ref{eq:reward_function}.

\begin{equation}
\label{eq:reward_function}
R=
\begin{cases}
    1.0, & \text{if }|x| < 50 \\
    c2^{-(|x|-\rho)/\tau)}, & \text{otherwise},
\end{cases}
\end{equation}

where the cliff constant $c=0.4$ ensures there is a distinct difference in the value of the state inside the target region compared to outside.

The MDP {action space} ($\mathcal{A}$) consists of three actions $a\in[a_0, a_1, a_2]$.  Representing the action space in this way decouples the ascent rate into desired altitude $a_0\in[14000, 21000]$\;km and time-factor $a_1\in[1, 5]$, as a component of time-steps, to reach that altitude.  A larger time-factor $a_1$ represents a slower ascent rate, as the agent decides lower urgency to reach the desired altitude. Decoupling the ascent rate augments the secondary function of resource conservation into the learning process.  From this perspective, altitude $a_0$ relates to wind speeds propelling the balloon to the target direction.  And the time-factor $a_1$ relates to the amount of resources used to reach that altitude.  

% Transparency
Another reason for decoupling the action space in this way is to incorporate transparency into the agent's decision making.  Compared to previous works on station-keeping, trajectories generated in the form of desired altitudes and time-factors are more transparent than low-level control inputs \cite{bellemare_autonomous_2020, xu_station-keeping_2022}.  Further work is needed to discern policy explanations from a generated trajectory in relation to the wind-field \cite{hayes_improving_2017}.

The third action $a_2\in[-1, 1]$ is the binary option to float vertically and takes precedence over the other two actions.  To discretise the action, the controller chooses to float if the action is positive, $a_2>0$.  In the event that action $a_2$ is selected, actions $a_0$ and $a_1$ become unnecessary and are subsequently disregarded.

An episode runs for a total of 3 days, where the balloon's \textbf{initial conditions} consists of the relative longitude, latitude and altitude from the target, date, time, and wind noise which are sampled uniformly from a random seed.  Every six hours, wind forecasts are obtained from ECMWF's ERA5 dataset \cite{hersbach_era5_2020}. To determine the wind speed at a specific time and position, multidimensional interpolation is applied to gather data from the wind field.  An episode ends when the agent reaches a flight time of 3 days or runs out of resources. 

Both the actor and critic networks are parameterised as neural networks, with two hidden layers of size $256$.  The actor network takes input $77$ which is the combination of the wind and ambient features and outputs a mean and covariance for each of the three actions, $\mu\in\mathbb{R}^{3}$, $\log\sigma\in\mathbb{R}^3$.  The critic network instead takes the concatenated vector of both the state and action and calculates the state-action value $Q(s_t, a_t)$.

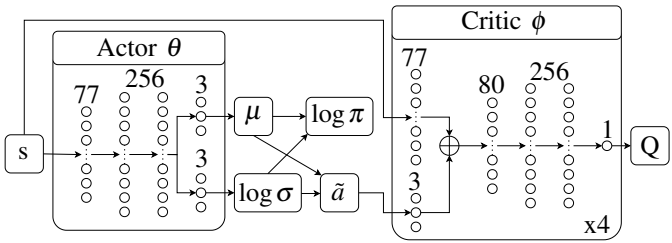
\begin{figure}[h]
\label{fig:network_architecture}
\centering

\begin{overpic}[]{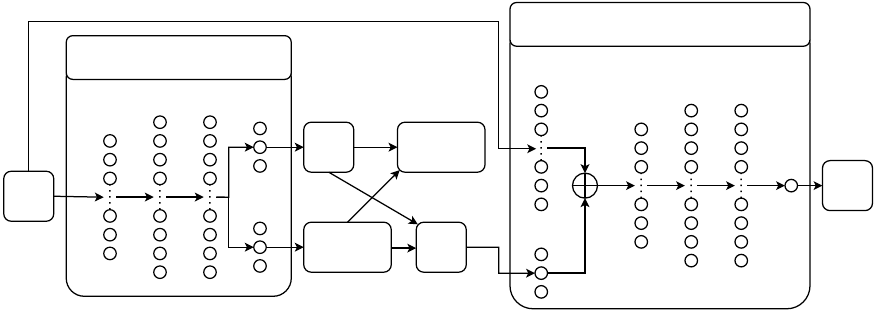}

\put(52, 76){\makebox(0,0){Actor $\theta$}}
\put(54, 64){\makebox(0,0){$256$}}
\put(8, 36){\makebox(0,0){s}}
\put(32, 58){\makebox(0,0){$77$}}

\put(75.5, 62){\makebox(0,0){$3$}}
\put(75.5, 34){\makebox(0,0){$3$}}

\put(100, 20){\makebox(0,0){$\log \sigma$}}
\put(127, 50){\makebox(0,0){$\log \pi$}}

\put(95, 50){\makebox(0,0){$\mu$}}
\put(127, 21){\makebox(0,0){$\tilde{a}$}}

\put(190, 85){\makebox(0,0){Critic $\phi$}}

\put(156, 25){\makebox(0,0){$3$}}
\put(156, 72){\makebox(0,0){$77$}}
\put(185, 62){\makebox(0,0){$80$}}
\put(207, 67){\makebox(0,0){$256$}}
\put(229, 45){\makebox(0,0){$1$}}
\put(245, 38){\makebox(0,0){Q}}
\put(225, 9){\makebox(0,0){x$4$}}

\end{overpic}

\caption{{\small Soft Actor-Critic \cite{haarnoja_off-policy_2018} network architecture.  The actor network is modelled as a Feed-forward parameterised Gaussian policy where the actions are represented as the hyperbolic tangent $\tanh$ applied to z values sampled from the mean $\mu$ and covariance given by the neural network.  Whereas the critic is modeled as a soft Q-function.}}
\label{fig:Pipeline}
\end{figure}

\subsection{Altitude Controller}

The altitude controller takes the desired altitude $a_0$, time-factor $a_1$, and float condition $a_2$ from the SAC controller, and either drops or vents helium depending on the calculated desired ascent rate.  Where the current altitude is denoted $h_t$ and the time between each action, stride time $\Delta t$, is set to 20 minutes.

\begin{equation}
    \dot{h}_{\text{des}} = \frac{a_0 - h_t}{a_1 \Delta t}
\end{equation}

%Feasibility
To prevent the agent from prioritizing receiving rewards by exploiting the environment \cite{amodei_concrete_2016}, a minimum limit for ballasting $m_{\text{s\_min}}$ and venting $n_{\text{min}}$ is set.  If ascent rates are chosen that calculate masses less than these values then no action is taken.  There is a loophole in this approach that could allow the agent to fly above or bellow the set altitude limits.  To prevent this behaviour, the controller will float above or bellow the altitude limits unless an action is chosen that propels the balloon between the altitude limits is chosen.  The full controller is shown in Eq \ref{eq:fullcontroller}. 

\begin{equation}
    \label{eq:fullcontroller}
    \text{action} = 
    \begin{cases}\text{Float,} & \text{if } a_2 > 0 \\
    \text{Do Nothing,} & \text{if } n_{\text{calc}} < n_{\text{min}} \text{ or } m_{\text{s\_calc}} < m_{\text{s\_min}}\\
    \text{Ballast,} & \dot{h}_{\text{des}} > \dot{h}_t \\
    \text{Vent,} & \dot{h}_{\text{des}} < \dot{h}_t \\
    \end{cases}
\end{equation}

Where $n_{\text{calc}}$ and $m_{\text{s\_calc}}$ are the number of mols and sand mass calculated to vent or ballast, respectively, for the given action.  The equations to calculate these are described below.

% \begin{equation}
% \frac{dh}{dt}_{desired}=
% \begin{cases}
% \frac{dh}{dt}_{desired}, & \text{if } |\frac{dh}{dt}_{desired}| > 0.5 \\
% 0.0 & \text{otherwise}\\
% \end{cases}
% \end{equation}

\subsection{Resources used to float, vent and ballast}

To calculate the desired mass to vent or ballast, we assume the balloon has reached steady state and equate Eq \ref{eq:fullN2L} to 0.  To \textbf{vent}, we solve Eq \ref{eq:fullN2L} for the number of mols $n_{\text{calc}}$ of helium given the desired ascent rate $\dot{h}_{\text{des}}$ from the SAC controller.  This results in a polynomial function which we can then solve for the real roots.

\begin{align}
\begin{split}
    \rho g \left(\frac{RT}{P} - M\right)n_{\text{calc}}
    - \frac{1}{2}\rho\left|\dot{h}\right|\dot{h}&C_d\pi\left(\frac{3RT}{4\pi P}\right)^{\frac{2}{3}}n_{\text{calc}}^\frac{2}{3} \\
    & -\left(m_p + m_s\right)g=0
\end{split}
\end{align}

Similarly, to \textbf{ballast} we can solve for the mass of sand $m_{\text{s\_calc}}$ by rearranging Eq \ref{eq:fullN2L}.

\begin{equation}
    m_{\text{s\_calc}} = \rho V - \frac{1}{2g} C_d A \left|\dot{h}\right|\dot{h} - m_p - m_h 
\end{equation}

When \textbf{floating}, the balloon has no vertical velocity, and hence drag is not acting on the balloon envelope.  This simplifies Eq \ref{eq:fullN2L} by setting the drag force $F_d$ equal to $0$.  Furthermore, we expand the mass term $m$ to include all the components.  This includes the mass of the payload (and envelope) $m_p$, sand $m_s$ and helium $nM_h$.  Where $n$ and $M_h$ are the number of moles and molar mass of helium respectively.

\begin{equation}
    0 = \rho V g - (m_p + m_s + nM_h)g
\end{equation}

Subsequently, we select the appropriate method to maintain equilibrium between the buoyancy force and weight of the balloon.  If the buoyancy force exceeds the weight of the balloon, we opt to vent helium $n_{\text{calc}}$.  Conversely, if the weight of the balloon is more dominant, we opt to ballast $m_{\text{s\_calc}}$:

\begin{equation}
\begin{cases}
    n_{\text{calc}} = \frac{m_s + m_p}{\frac{\rho R T}{P}-M_h}, & \text{if } \rho gV > mg \\
    m_{\text{s\_calc}} = \rho V - m_p - m_h, & \text{otherwise}\\
\end{cases}
\end{equation}

%% file: 5_Results.tex
\section{Results}

\subsection{Equations of Motion}
%Evaluating the solver calculates the correct number of mols.

\begin{figure}[b]
    \centering
\begin{overpic}[abs, width=\linewidth,unit=1mm,scale=.25]{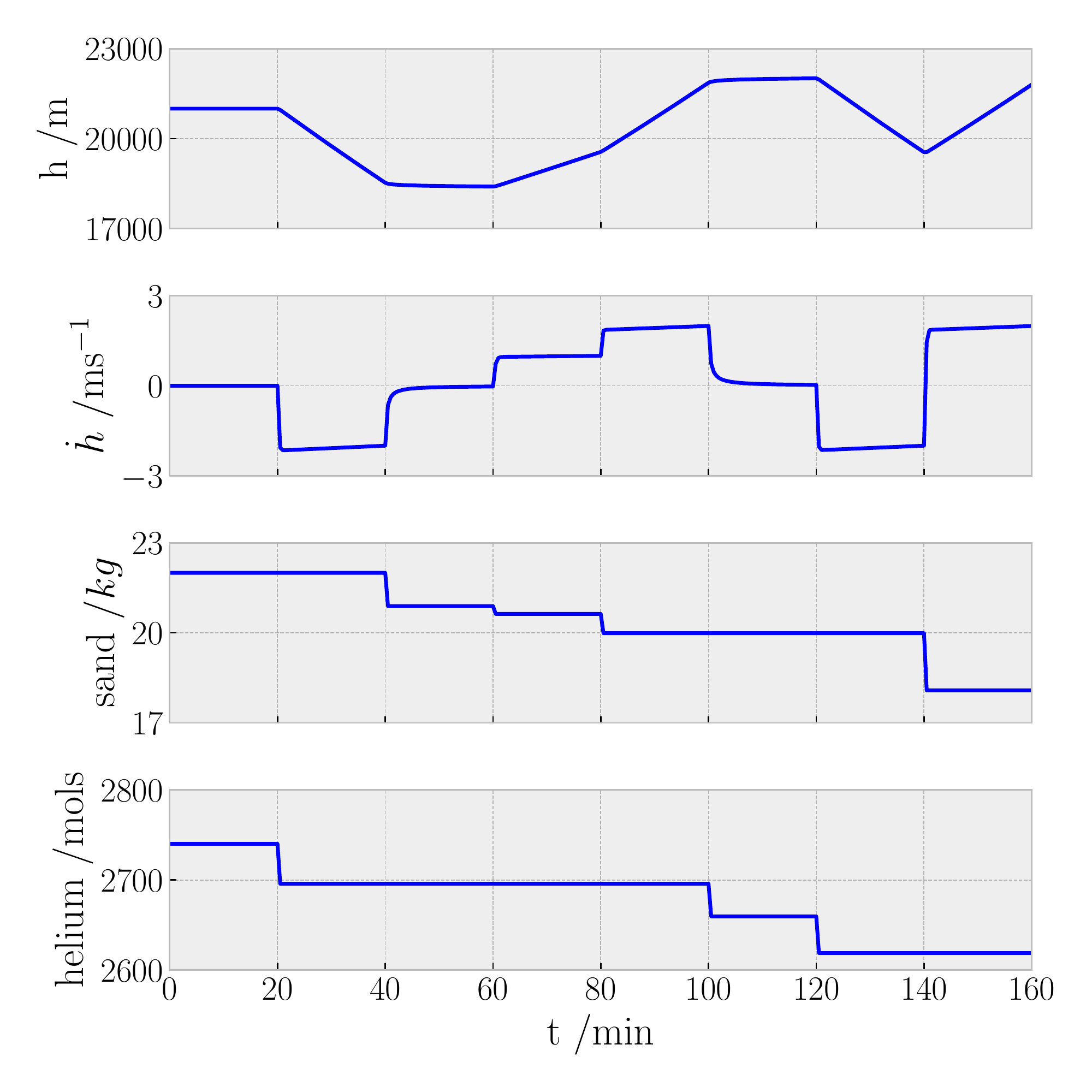}
\end{overpic}
\vspace*{-8mm}
    \caption{{\small The impact of venting and ballasting on altitude and ascent rate, with discrete actions applied at 20-minute intervals.}}
    \label{fig:AltitudeControllerTest}
\end{figure}

The altitude calculations and resources used are validated in simulation. Through a set of venting, ballasting and float commands, the desired ascent rate matches the ascent calculated from the solver in Fig \ref{fig:AltitudeControllerTest}.

%Modelling venting and ballasting over time instead of instantaneous change
Of note is the instantaneous change in ascent rate that follows from either venting or ballasting.  Modelling the mass flow rate of the valve and ballast would lead to a more accurate estimation of the state of the balloon.  %\todo{what effect do this have on the balloon w.r.t it's altitude, velocity?}

%Not modeling the pressure difference between the ambient air and balloon leading to a slight variation in the ascent rate (where we assume it is constant)
The ascent rate of the balloon can also be seen to increase gradually. Although an assumption made is that the ascent rate is constant, we believe this effect is a result of the assumed property of the internal gas equalling the pressure of the ambient air.  Further modelling of the stress profile of the latex is required to model the pressure difference between the internal pressure and ambient pressure at varying altitudes. As is evident from Fig \ref{fig:AltitudeControllerTest}, this effect is minor.

\subsection{Station Keeping Performance}

\begin{figure}[h]
    \includegraphics[width=1.05\linewidth]{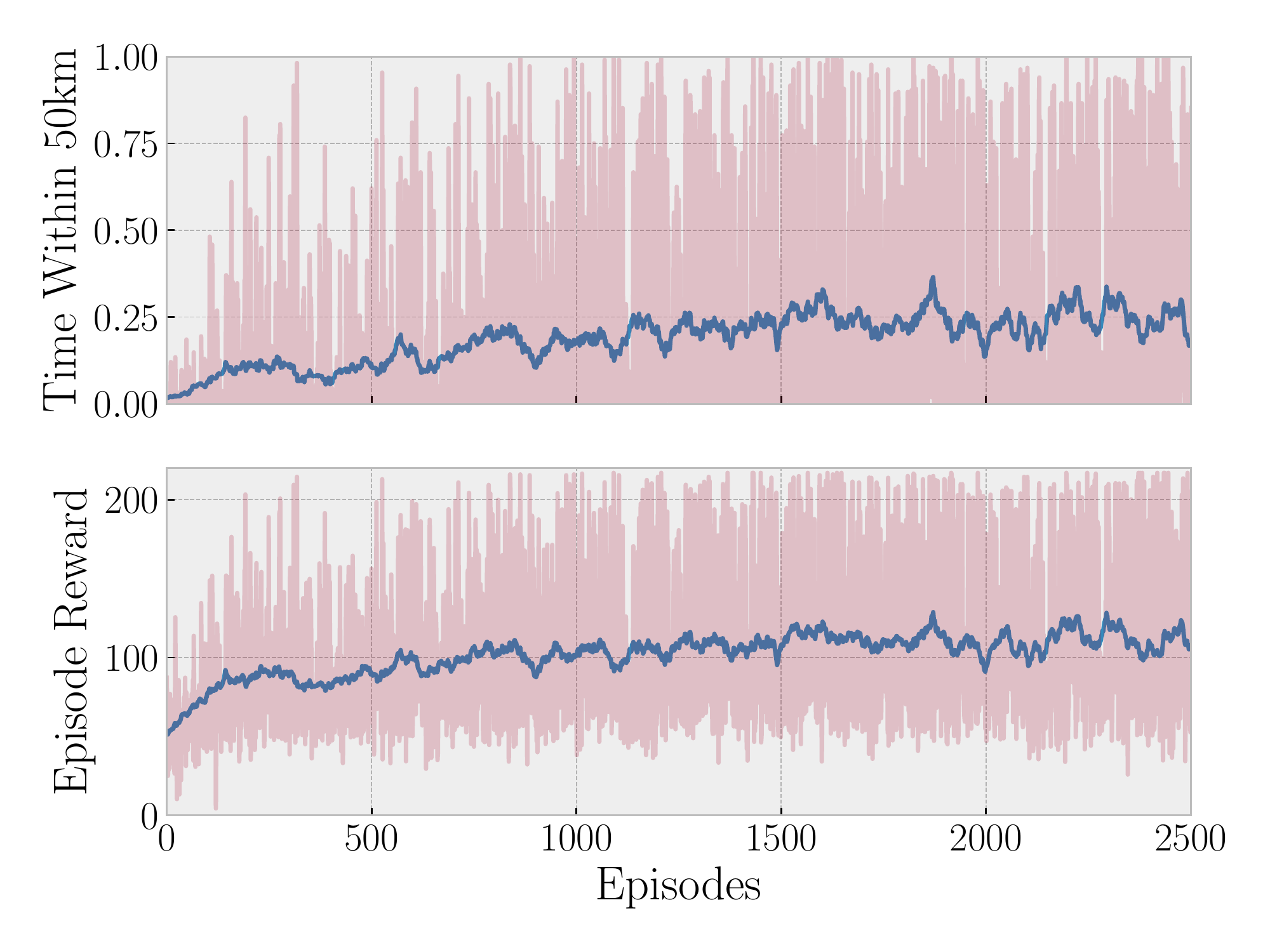}
    \vspace*{-8mm}
    \caption{{\small The cumulative episode reward and success metric time within 50km for each episode.  Training lasted for 3 hours where the controller was trained for a total of 2500 episodes.  The blue line represents the rolling average over 50 episodes.}}
    \label{fig:tw50_reward}
\end{figure}

%Graph - Just saying what's there
The training performance of the SAC controller, as shown in Fig \ref{fig:tw50_reward}, achieves an average episode reward of 110 and time within 50km (TW50) of 25\%.  The metric TW50 represents the percentage of the total episode length spent within a radius of 50km of the target.
%Training process
The results indicate that the agent is capable of achieving a moderate level of proficiency in station keeping.  Yet achieving a proficient level requires a prolonged training period.

\begin{figure}[h]
    \includegraphics[width=1.05\linewidth]{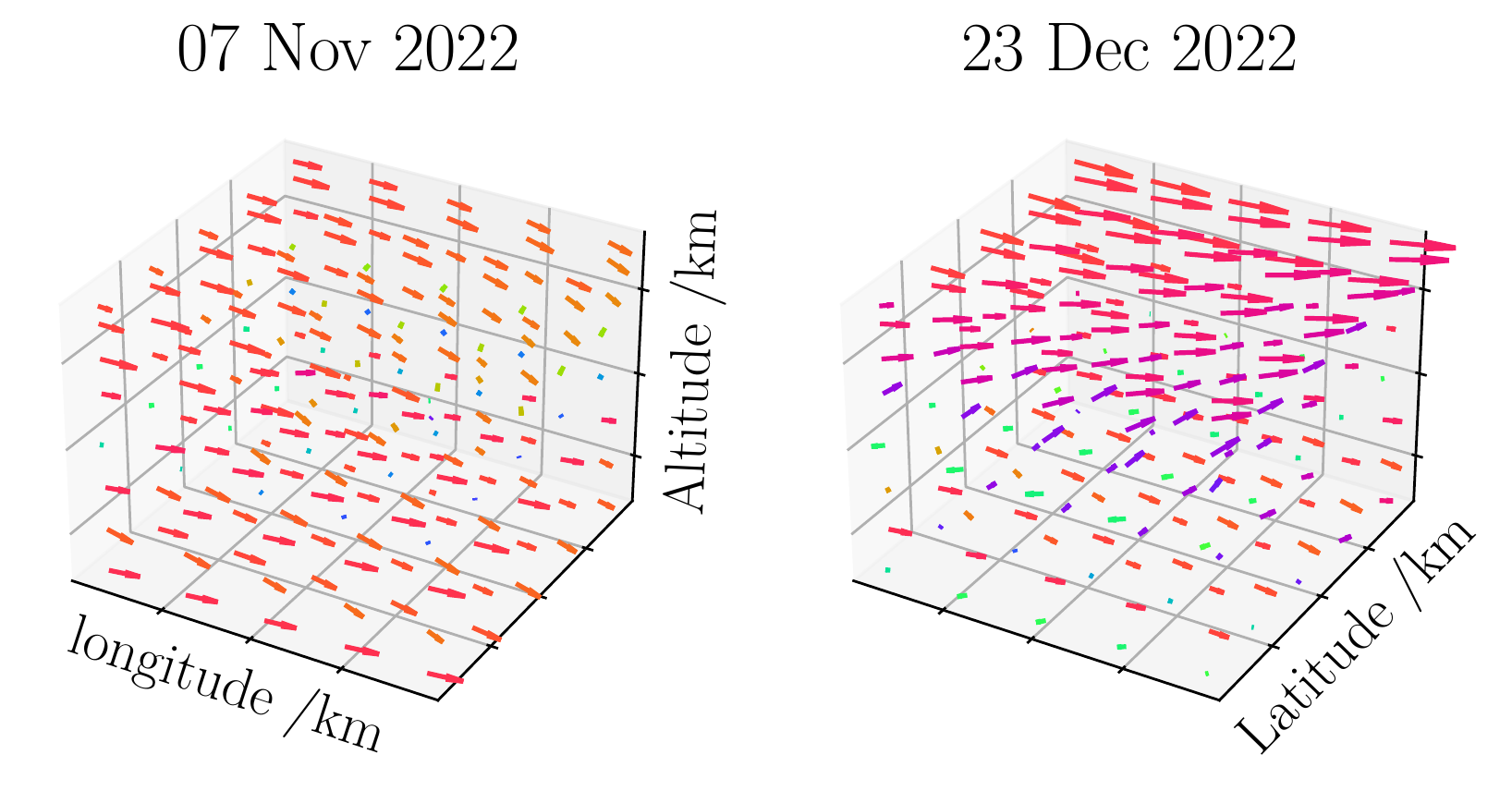}
\caption{{\small Comparison between wind fields which are less (7th of November 2022) and more (23rd of December 2022) favorable to navigate.  This is due to the variation in wind directions indicated by the color of the arrow.}}
\label{fig:differingwindfields}
\end{figure}

% difficult to navigate wind fields (low variety, only going in one direction)
The controller clearly exhibits suboptimal performance for a portion of the wind fields.  Upon closer analysis of these fields, it becomes apparent that there is a subset of wind fields in which the wind direction is either uniformly oriented, illustrated in Fig \ref{fig:differingwindfields}, or contains exceptionally high wind speeds.  
%Uniform oriented wind fields stopping station keeping
The presence of uniformly oriented wind directions can pose challenges when attempting to station keep, as the balloon cannot leverage opposing wind directions to station keep.  
% High wind speeds cause quick deviation, low action time not enough to stop this
Furthermore, high wind speeds can cause the balloon to quickly deviate from the intended path.  Potentially due to the fact that the controller action time of 20 minutes may not be sufficient to counteract the strong wind forces.

\begin{figure}[t]
    \centering
    \vspace*{8mm}
    \includegraphics[width=\linewidth]{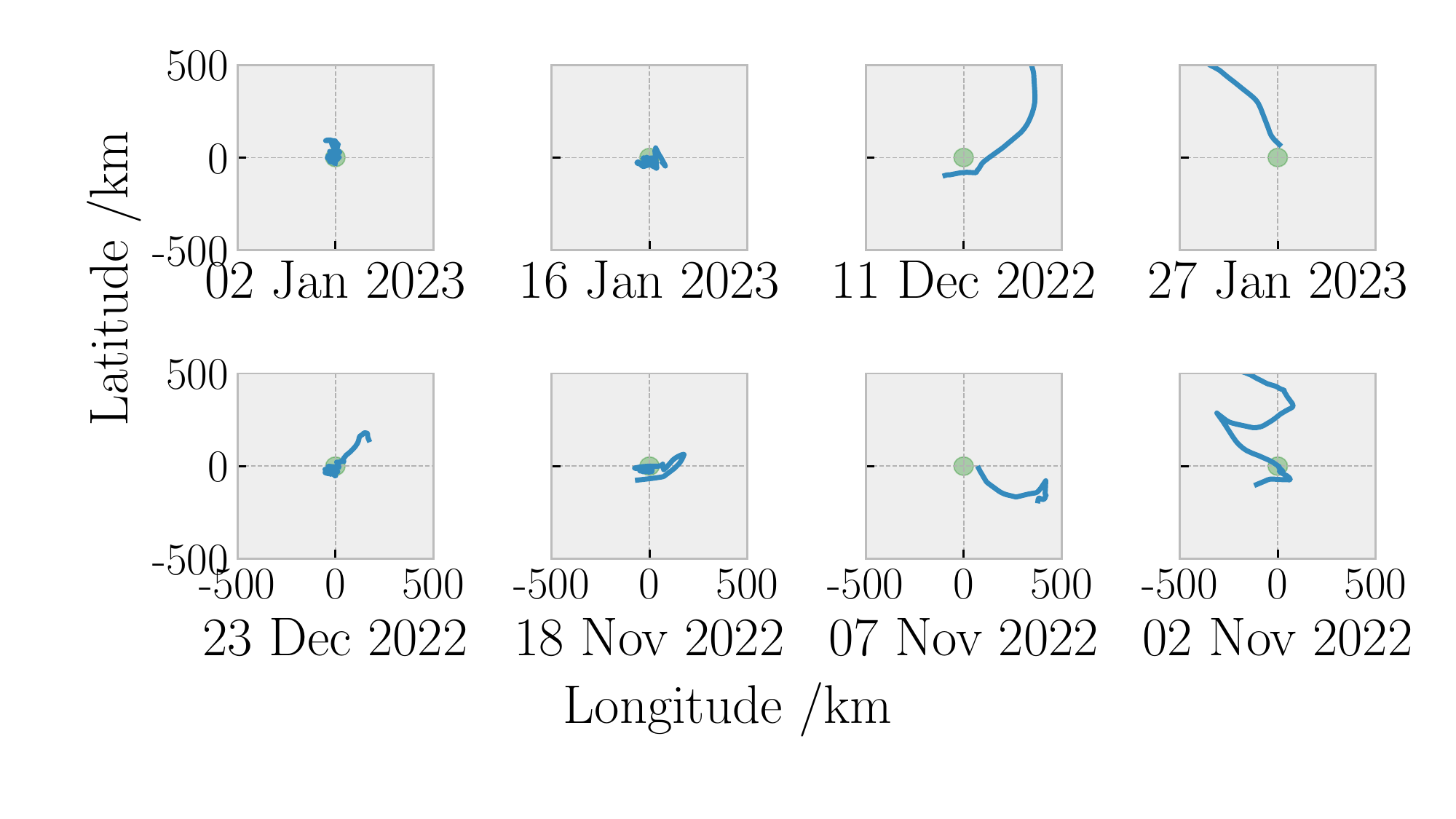}
    \vspace*{-10mm}
    \caption{{\small Illustration of good and poor station-keeping performance.  Where good performance is illustrated by the length of time spent within the target region denoted by the green circle.}}
    \label{fig:birdseye_view}
\end{figure}

% Birds eye view - good station keeping staying within the region
% Bad station keeping due to wind field characteristics
% Characteristic 1: no wind vector pointing towards station
% Characteristic 2: no diverse wind vectors to station keep
Fig \ref{fig:birdseye_view} illustrates the performance of the controller from a birds-eye view across various days. The wind-field for the 23rd of December, shown in Fig \ref{fig:differingwindfields}, and 18th of November demonstrate the controller's capability to station-keep, with a high duration of time spent within the target region. However, the wind-fields on the days of the 2nd and 7th of November emphasize the close relationship between the controller's performance and the characteristics of the wind-field.  For the 7th of November also shown in Fig \ref{fig:differingwindfields}, the balloon was unable to find wind directed towards the target.  Whereas on the 2nd of November, the controller failed to find a set of diverse fields to station keep at the target. These observations necessitate further exploration of the wind-field, which is challenging as the controller can  access only the wind-vectors at the balloon's current location due to partial observability of the state.

\subsection{Conservation of Resources}
\begin{figure}[h]
    \centering
    \includegraphics[width=\linewidth]{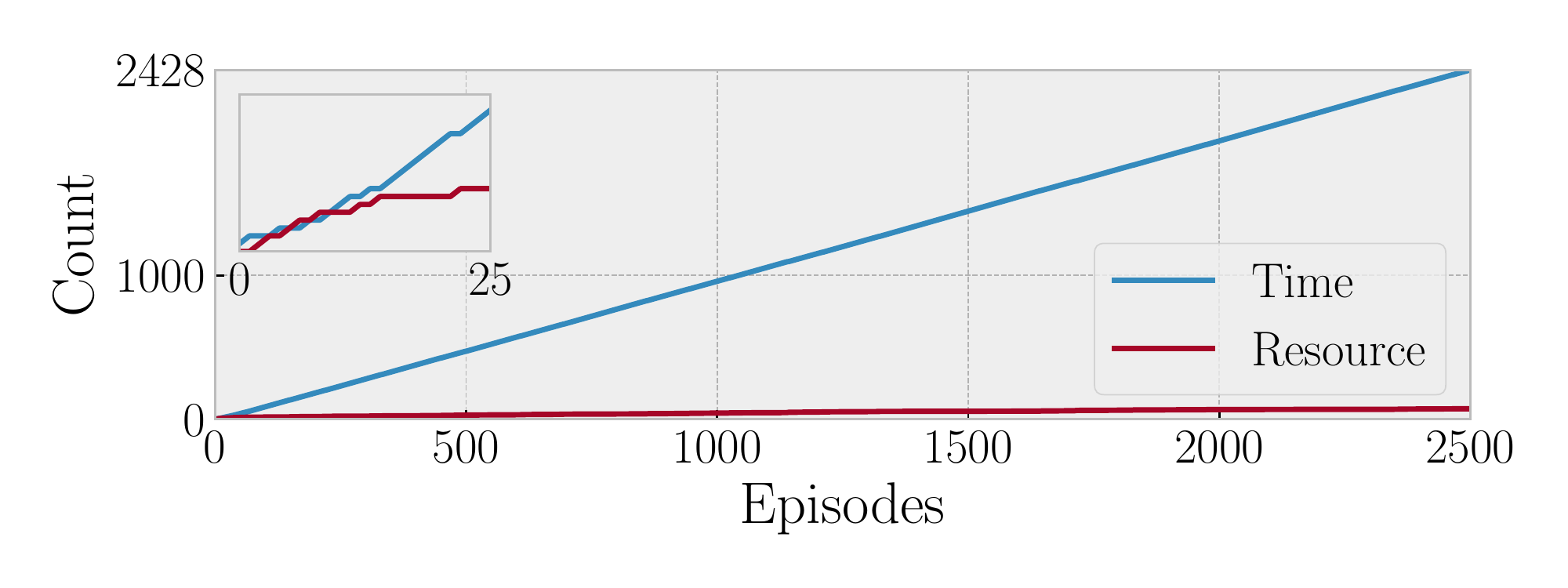}
    \vspace*{-7mm}
    \caption{{\small The cumulative terminal conditions for every episode. 
 The agent learnt very quickly after only a few episodes to conserve resources.  The terminal condition Time represents a flight lasted 3 days, whereas resources represents an episode where the balloon ran out of sand to ballast.}}
 \label{fig:ResultsTW50TerminalResources}
\end{figure}

% Conserve resources - high ascent rates lead to early episode termination, then doesn't ustalise all it's resuources early on
Finally, we evaluate the resource conservation behavior of the RL controller to prevent premature termination of episodes. Figure \ref{fig:ResultsTW50TerminalResources} shows how the controller learns to conserve resources within 25 episodes.  Afterwards, the agent rarely ends the episode prematurely and learns not to utilise all its resources early on.

\begin{figure}[h]
    \centering
    \includegraphics[width=\linewidth]{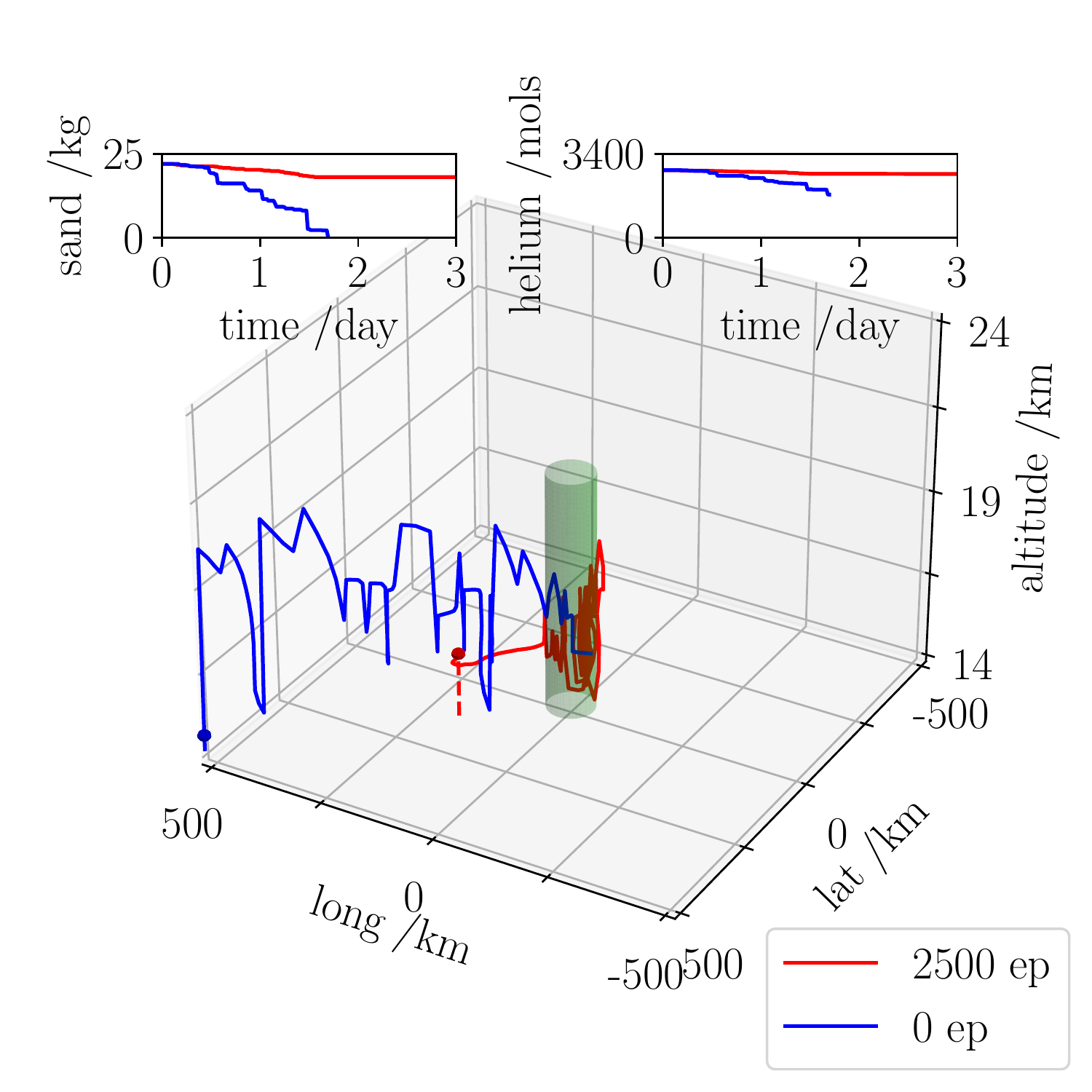}
    \vspace*{-6mm}
    \caption{{\small Illustration of an untrained agent compared to an agent trained for 2500 episodes with the same initial conditions for the wind-field on the 23rd December.  The untrained agent utilises all of it's resources quickly which leads to an early episode termination. Alternatively, the trained agent uses resources conservatively, while also achieving a high station keeping performance.}}
    \label{fig:3d_plt_2022_12_23}
\end{figure}

This is even more evident when an agent trained for 2500 episodes is compared against an agent not trained.  As shown in Fig. \ref{fig:3d_plt_2022_12_23}, the trained agent is capable of using winds at alternative altitudes to stay within the station-keeping radius, indicated by the green cylinder.  Whereas the untrained agent, is propelled by the wind currents away from the target while also using all of its resources leading to an early episode termination.

%% file: 6_Conclusion.tex
\section{Conclusion}

In this paper, to the best of our knowledge, we introduce the first-ever station-keeping controller using deep reinforcement learning for a helium-based balloon actuated through the venting of helium and ballasting of sand. 

% Results TW50km and conserving resources
We show through experiments using real wind-field forecasts from the ERMWF ERA5 dataset that a reinforcement learning-based controller, Soft Actor-Critic, can achieve 25\% time within 50km while also conserving helium and sand to endure a 3 day flight.  We train our controller between the dates of November 2022 and January 2023 within the tropics at longitude -113$^{\circ}$ latitude 1$^{\circ}$.
% Resource usage
The proposed controller maximizes the duration of the balloon's position within the target region while also effectively minimising the consumption of resources, thereby supporting long duration flights.
% Realistic resource usage
Furthermore, using the derived equations of motion for the balloon, realistic actions can be achieved by setting a minimum venting and ballasting actions.

% SAC utalised including entropy, allows to explore the wind field 
The incorporation of augmented reward with policy entropy in the Soft Actor-Critic (SAC) algorithm is shown as a promising approach for enabling effective exploration of the wind-field to identify desirable wind-vectors directed towards the target, as well as diverse wind-vectors that are beneficial for station-keeping.  
% Continuous action spaces
Furthermore, using continuous action spaces enables us to exploit a wider range of ascent rates, which are not attainable through the application of discrete action spaces, as has been commonly employed in previous studies. This enhancement in the action space allows for a more finely-grained control of the ascent rate and enables the controller to make more precise adjustments to the balloon's altitude.  
% Transparency, the action space is more transparent than previous works,
The decomposed desired ascent rate into desired altitude and time-factor also aids in the transparency of the behaviour of the agent given that the trajectories generated in the form desired altitudes and time-factors are more transparent than low-level control inputs.
%Future work
For future work, we intend to test the simulated controller on a real balloon to evaluate its station-keeping performance and to analyse the sim-to-reality gap.